\documentclass[lettersize,journal]{IEEEtran}
\usepackage{amsmath,amsfonts}
\usepackage{algorithmic}
\usepackage{algorithm}
\usepackage{array}
\usepackage{flushend}
\usepackage{textcomp}
\usepackage{stfloats}
\usepackage{url}
\usepackage{verbatim}
\usepackage{graphicx}
\usepackage{subcaption}
\usepackage{pgf}
\usepackage{cite}
\usepackage{booktabs}
\usepackage{todonotes}
\usepackage{bm}
\usepackage{url}
\usepackage{hyperref}
\usepackage{siunitx}
\usepackage{array}
\usepackage{graphicx}
\usepackage{amsthm}
\usepackage{siunitx}
\usepackage{nicefrac}
\newtheorem{assumption}{Assumption}

\newcolumntype{L}[1]{>{\raggedright\arraybackslash}p{#1}}

\newcommand\mat[1]{\bm{#1}}
\renewcommand\vec[1]{\bm{#1}}
\newcommand\fk{\Phi}

\definecolor{OliveGreen}{rgb}{0,0.6,0}
\begin{document}
\sloppy
\title{MUKCa: Accurate and Affordable Cobot Calibration Without External Measurement Devices}

\author{Giovanni Franzese*, Max Spahn*, Jens Kober,
and Cosimo Della Santina
\thanks{* Denotes equal contribution}}
\maketitle
\begin{abstract}

To increase the reliability of collaborative robots in performing daily tasks, we require them to be accurate and not only repeatable.
However, having a calibrated kinematics model is regrettably a luxury, as available calibration tools are usually more expensive than the robots themselves.  
With this work, we aim to contribute to the democratization of co-bots calibration by providing an inexpensive yet highly effective alternative to existing tools.
The proposed minimalist calibration routine relies on a 3D-printable tool as only physical aid to the calibration process. This two-socket spherical-joint tool kinematically constrains the robot at the end effector while collecting the training set. An optimization routine updates the nominal model to ensure a consistent prediction for each socket and the undistorted mean distance between them. We validated the algorithm on three robotic platforms: Franka, Kuka, and Kinova Cobots. 
The calibrated models reduce the mean absolute error from the order of \SI{10}{mm} to \SI{0.2}{mm} for both Franka and Kuka robots.  
We provide two additional experimental campaigns with the Franka Robot to render the improvements more tangible. First, we implement Cartesian control with and without the calibrated model and use it to perform a standard peg-in-the-hole task with a tolerance of \SI{0.4}{mm} between the peg and the hole. Second, we perform a repeated drawing task combining Cartesian control with learning from demonstration. Both tasks consistently failed when the model was not calibrated, while they consistently succeeded after calibration.
\end{abstract}

% \begin{IEEEkeywords}
% Kinematic Calibration, Cobots, Affordable hardware, Fine Manipulation
% \end{IEEEkeywords}

\section{Introduction}
To face the challenge of modern manipulation tasks in unstructured environments, where robots need to adapt their behavior in the Cartesian space, we require robots that are not only repeatable but also accurate. However, due to manufacturing and assembly tolerances, wears, and bending, errors occur over the entirety of the kinematic chain.

Kinematic errors are especially harmful in serial manipulators where the parameter deviations propagate through the kinematic chain resulting in wrong end-effector prediction. 
This is a known and well-studied problem \cite{xuan2014review}, and many calibration routines for industrial robots have been proposed. However, many solutions require the use of precision tools, like laser trackers or precision probes, which have different drawbacks: they are not portable, require an expert to set up, and are very expensive. For this reason, when working with affordable collaborative robots, the calibration of the kinematics is not considered worth the investment. 

\begin{figure}
    \centering
    \vspace{-15mm}
    \includegraphics[width=1\linewidth]{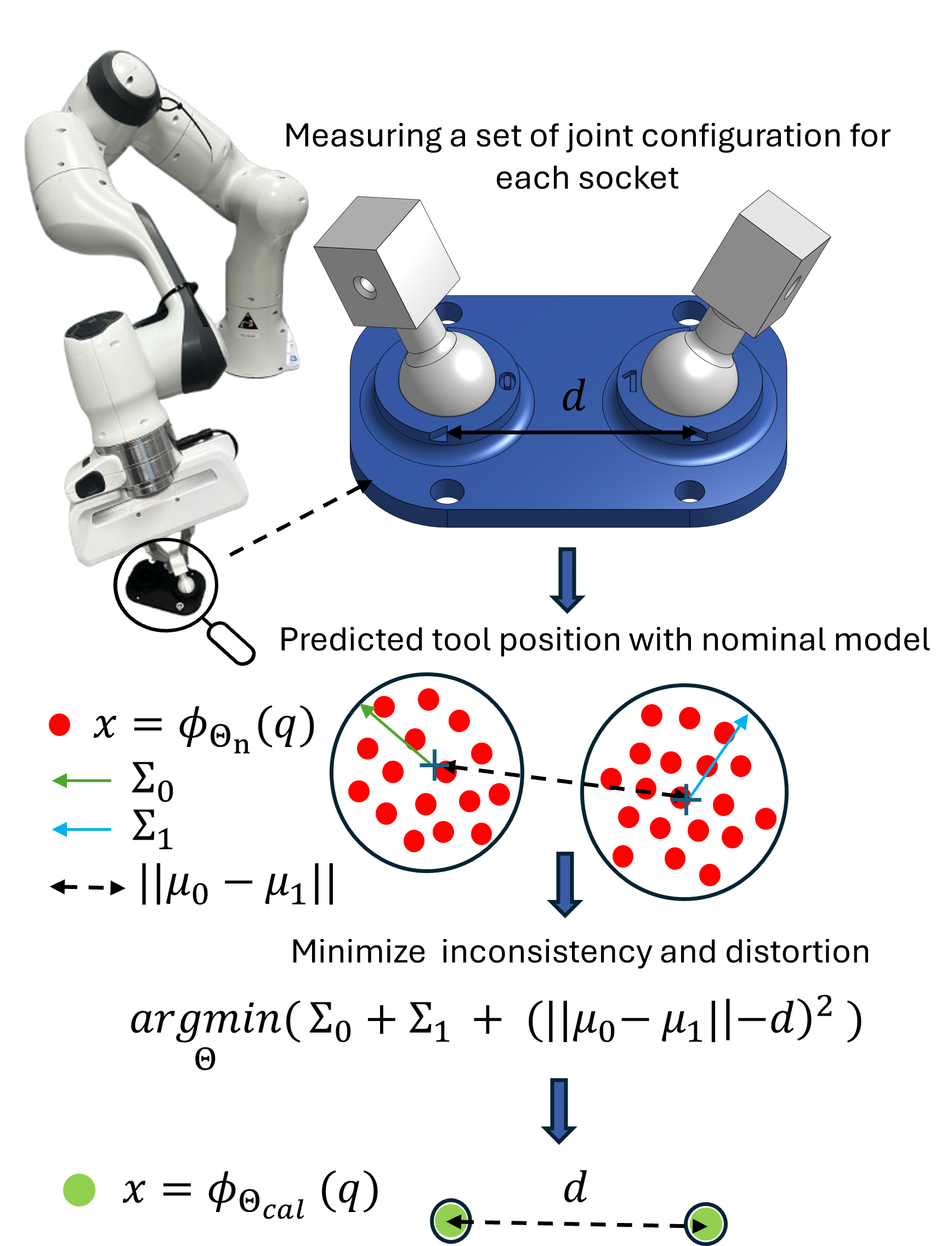}
    \caption{Calibration scheme: the optimization minimizes inconsistency and distortion of the predicted ball position attached at the end-effector.}
    \label{fig:cover}
\end{figure}

This article's contribution is the  Minimalist and User-friendly Kinematics Calibration (MUKCa) as the combination of an affordable calibration tool, illustrated in Fig. \ref{fig:cover} and
an optimization algorithm for kinematic parameter identification. The proposed minimalist tool is designed for accessibility and affordability, and it is 3D printable and composed of a sphere (or ball) with a two-socket base, as illustrated in Fig. \ref{fig:cover} without relying on external measurement systems. After printing them, the calibration can immediately start with the data recording on the robot by simply placing the tool in front of the robot and the sphere attached to the end-effector. 

The two sockets are at a known distance, and the ball has minimal play in the sockets. 
After collecting set of joint configurations for each socket, the calibration routine minimizes the variance of the predicted ball position for each of the two sockets, maximizing the model nullspace \textbf{consistency}.  Moreover, the model is optimized such that the distance among the average prediction matches the known distance between the two sockets, minimizing the model volumetric \textbf{distortion}. This consistency-and-distortion optimization is the first of its kind in the field of kinematic calibration. 
The next section will highlight and compare the most relevant works that also aim to create portable and affordable calibration routines. 

\section{Related Works}
The selection of measurement tools for kinematics calibration often involves trade-offs between cost, precision, and portability. The calibration
methods with external measurement devices are called the \textbf{direct} or \textbf{open loop} method\cite{xuan2014review}. 
Common measurement tools include Coordinate-Measuring Machines (CMMs) \cite{nubiola2014comparison} for high precision but low portability, laser trackers \cite{li2016poe, nubiola2013absolute} for real-time tracking at high cost, and Motion Capture \cite{pascal2023ros} or standard cameras \cite{feng2025novel} for concept validation but limited precision.   

A different family of calibration procedures does not rely on measuring the ground truth position (or pose) of the end effector. For this reason, they are referred to by the name of \textbf{indirect} or \textbf{closed loop} methods. 
The most common setup is the three metal sphere artifacts \cite{joubair2015kinematic, gaudreault2018self, nadeau2019impedance}.  
The main idea is that the metal spheres are manufactured to be perfectly spherical and are placed at a known precise distance from each other. The spherical constraints of the metallic surface and the sphere's radius are used to update the kinematic model. However, the points on the surface of each sphere are recorded using a precision-probe \cite{joubair2015kinematic} to increase accuracy, making the setup not affordable to every pocket. 
Additionally, the data collection often requires more than 30 minutes. 
In order to remove the use of the precision probe, three digital indicators, fixed orthogonally to each other, are proposed in \cite{gaudreault2018self}. Rather than reconstructing the surface of the metal ball, the digital indicator is used to automatically align the robot TCP with the center of each of the three balls on the plate. Despite the portability of the tool, precision sensors are still required to perform the automatic alignment, which keeps the cost prohibitive. 
To avoid the usage of expensive measuring devices, Nadeau et al. \cite{nadeau2019impedance} proposed a kinematic coupling device, i.e., a magnetic socket attached at the end effector that perfectly fits each of the three spheres. The kinematic calibration is then performed by solving an optimization problem where the fitting cost measures the error of the estimated distances between the sphere and the real one.
Since the achieved accuracy relies directly on the tool's accuracy, its calibration faces similar limitations as the previously mentioned methods, particularly in terms of cost and portability.

Recently, Petriv et al. \cite{petrivc2023kinematic} highlighted the use of spherical joints, as a cost-effective alternative to the three-sphere system. The proposed method is to measure different configurations with an end-effector ball in different sockets placed in the workspace. However, since the sockets are hand-positioned, it is not possible to know the accurate distance between them. The proposed fitting function is the minimization of the discrepancy of the predicted forward kinematic position of each socket with respect to the average value of the nominal model. The use of the average nominal model as ground truth is a strong limitation that our proposed method gets rid of. 

As highlighted in the next section, our proposed method combines the cost-effectiveness of spherical joints while:

\begin{itemize}
    \item eliminating the limitation of relying on the average nominal model \cite{petrivc2023kinematic},
    \item avoiding the need for an expensive CMM to calibrate the distance between the sockets \cite{gaudreault2018self} or using an external measure to define the calibration tool frame, and
    \item removing the requirement for a costly auto-centering device for the spherical joint in the socket \cite{nadeau2019impedance}.
\end{itemize}

Instead, we propose a low-cost alternative where the user kinesthetically positions the ball within the socket for different joint configurations. 
This compromise provides the best balance between affordability, accuracy, and practicality.

\section{Kinematics Description and Optimization}
\label{sec:methodology}

We first describe the chosen forward kinematics formulation and introduce
its parametrization. Then the proposed optimization formulation as the 
calibration method is formulated.

\subsection{Kinematics}

A kinematic chain is composed of $m$ links and $n$ joints, where $m \geq n$. A transformation between
two frames $a$ and $b$ on the kinematic chain is described by a
homogeneous transformation matrix $\mat{T}_a^b$. The frame $0$ represents the robot's base
frame and some frames are referred to by a descriptive name, such as \textit{sphere} or \textit{flange}. We refer to $\vec{X}_a^b$ as the translational part of the transformation, in other words the last column of the matrix excluding the 4-th row.
There are generally two types of joints: (a) fixed joints where the corresponding transformation matrix
is purely dependent on the geometry and (b) actuated joints where the transformation matrix is additionally
dependent on the joint value. 

The transformation matrix for a fixed
joint between frame $i-1$ and $i$ can
be obtained from the Roll-Pitch-Yaw angles $[\alpha,\beta,\gamma]_{i-1}$ and
the displacement vector
$\vec{p}_{i-1} = [p_x, p_y, p_z]_{i-1}$ as:
\[
   \mathbf{T}^i_{i-1} =
   \begin{bmatrix}
   \mat{R}([\alpha,\beta,\gamma]_{i-1}) & \vec{p}_{i-1} \\
   \mat{0}^\top & 1
   \end{bmatrix},
\]

For a revolute joint define by an axis of rotation $\vec{a}_i$ and the angle $q_i$, the
transformation is multiplied from the right with an additional rotation, namely:
\[
   \mat{T}^i_{i-1} =
   \begin{bmatrix}
   \mat{R}([\alpha,\beta,\gamma]_{i-1}) \cdot \mat{R}(\vec{a}_i, q_i) & \vec{p}_{i-1} \\
   \vec{0}^\top & 1
   \end{bmatrix},
\]
where 
\[
    \mat{R}(\vec{a}_i,q_i) = \mat{I} + \sin q_i \, \mat{K} + (1 - \cos q_i) \, \mat{K}^2,
\]
and the skew-matrix $\mat{K}$ is given by
\[
K =
\begin{bmatrix}
0 & -a_{i,2} & a_{i,1} \\
a_{i,2} & 0 & -a_{i,0} \\
-a_{i,1} & a_{i,0} & 0
\end{bmatrix}.
\]

For other type of joints, e.g., prismatic, similar formulas can be found.
The last frame on the kinematic chain, frame $m$ is then obtained by the product
of transformations 
\[
 \mat{T}_0^m = \mat{T}_0^1\mat{T}_1^2\cdot\mat{T}_{m-1}^m.
\]

In our approach, we use kinematic models that are parameterized by the Euler
angles $[\alpha,\beta,\gamma]_i $  and the displacements
$\vec{p}_i \  \forall i=0,...,m$. This is a deliberate choice over
other parametrizations, such as Denavit-Hartenberg parameters \cite{petrivc2023kinematic}
or the Product of Exponentials \cite{li2016poe}, Vector Inner Product Error Model\cite{liu2024kinematic} or Linear Finite Screw Deviation Model \cite{kim2024serial} because most robot models
are available in the Universal Robot Description Format (URDF). To favor 
simplicity of the method over minimalistic parametrization, we avoided the
conversion between the representations \cite{huczala2022automated}.
Finally, we concatenate all parameters into one parameter vector $\vec{\Theta} \in \mathcal{R}^{6m}$, such that we can write the parameterized forward kinematics as a function of the parameters $\vec{\Theta}$ and the joint angles $\vec{q}$, i.e.,  $\mat{T}_0^m(\Theta, \vec{q})$. 

\subsection{Data collection}
We record two sets of configurations, one with the sphere in the socket ``zero" and one in the socket ``one". Figure~\ref{fig:robots} depicts the users that kinesthetically move the robot in the nullspace and record the configuration that ends with the sphere in one of the two sockets. 
This is a key component of the user-friendliness of the proposed methodology that does not require designing an ad hock controller or planning strategy to satisfy that constraint. 
This  brings the
\begin{assumption}
    The robot's motors are back-drivable. 
\end{assumption}

We identify the two sets with the corresponding apex, i.e. $ \bm{q}^{k} \in \mathbb{R}^{1 \times m}$ where $m$ is the number of joints in the kinematic chain and $k$ is the index of the socket where the kinematic chain terminates. Given a set of kinematic parameters, the predicted \textbf{ball center point} (BCP) for each set are 
$$\bm{X}^{0}=\fk_{\vec{\Theta}}(\bm{q}^{0}) \text{   and   } \bm{X}^{1}=\fk_{\vec{\Theta}}(\bm{q}^{1})$$
where $\fk_{\vec{\Theta}}$ is the parameterized forward kinematics of the BCP.
However, the necessity to record a set of  configurations that have the BCP in the same sockets implies the 
\begin{assumption}
    The robot is kinematically redundant, meaning it can reach the same end-effector position with different joint configurations.
\end{assumption}
The violation of this assumption would make the recording of a set of measurements for each socket impossible. 

Moreover, the uncertainty on the predicted Cartesian position is given by the uncertainty on the robot parameters $\Delta \vec{\Theta}$ (due to the manufacturing and assembly process) and on the joint angle $\Delta \bm{q}$ (due to noise in the encoder measurements or on the unobservable gear backlash and shaft flexibility). Nevertheless, we hypothesize the 
\begin{assumption}
\label{assumption:no_joint_noise}
    The uncertainty on the joint state \( \bm{q} \) is negligible, i.e. \( \Delta \bm{q} \approx 0 \).
\end{assumption}
This implies that the observed Cartesian error $\Delta \bm{x}$ is only due to miscalibrated geometry modeling $\Delta \vec{\Theta}$. This is a fair assumption when using collaborative robots that use harmonic drive gearing that is known to have negligible backlash. However, this assumption is usually overlooked in other kinematic calibration methods but is a necessary condition to guarantee the convergence of the optimization. 

\subsection{Parameter optimization}
Given a vector of parameters $\vec{\Theta}_i$ and $N^{(0,1)}$ measured joint configurations for each socket, the mean square error with respect to the ground truth Cartesian position is: 
\begin{equation}
   J_{\text{mse}} = \frac{1}{N}\sum_{i=1}^N \lVert\fk_{\vec{\Theta}}(\bm{q}^i) -\vec{X}^i\rVert_2^2 
\end{equation}
where $N = N^{(0)}+ N^{(1)}$. 
Moreover, we know that the BCP is located in only two positions. We can rewrite the MSE as
\begin{equation}
   % J_{\text{mse}} = 
   \small
   \frac{1}{N}\left ( \sum_{i=1}^{N^{(0)}} \lVert\fk_{\vec{\Theta}}(\bm{q}^{(0)}) -\bm{X}^{(0)}\rVert_2^2 + \sum_{i=1}^{N^{(1)}} \lVert\fk_{\vec{\Theta}}(\bm{q}^{(1)}) -\bm{X}^{(1)}\rVert_2^2 \right )
\end{equation}
that can be rewritten as the weighted sum of the mean squared error with respect to each of the holes, i.e., 
\begin{equation}
   J_{\text{mse}} = 
   \small
   \frac{1}{N}\left ( {N^{(0)}} J^{(0)}_{\text{mse}} + {N^{(1)}} J^{(1)}_{\text{mse}} \right )
\end{equation}
By using the variance-bias decomposition of each of the two mean squared error costs, we obtain 
\begin{equation}
    J^{(0)}_{\text{mse}} = \underbrace{\mathrm{tr}(\bm{\Sigma}_{(0)})}_{variance} +  \underbrace{||\bm{\mu}_{(0)} - \bm{X}^{(0)}||_2^2}_{bias}
\end{equation}
where $N = N^{(0)} + N^{(1)}$ and the mean(s) are defined as  
\[
   \bm{\mu}^{(0,1)} = \frac{1}{N^{(0,1)}} \sum_{i=1}^{N^{(0,1)}} X_i^{(0,1)},
\]
with $\bm{\mu} \in \mathbb{R}^3$, 
and the variance matrices are defined as
\[
   \bm{\Sigma}^{(0,1)} = \frac{1}{N^{(0,1)}} \left(\bm{X}^{(0,1)} - \bm{\mu}^{(0,1)} \right) ^\top \left(\bm{X}^{(0,1)} - \bm{\mu}^{(0,1)} \right)
\]
with $\bm{\Sigma} \in \mathbb{R}^{3 \times 3}$. However, the calculation of the bias requires the measurement of the absolute position of the socket, which is not readily available without relying on an expensive tracking system.
 
Nevertheless, we do have a ground truth measure, which is the distance between the two sockets.

To ensure the distance preservation of the final model, we can minimize the bias in the estimation of distance between the sockets, i.e.,
$$
J_{\text{bias}}=(\lVert \bm{\mu}_{(1)}- \bm{\mu}_{(0)}  \rVert _2 - {d})^2.
$$

Hence the proposed cost function becomes
\begin{equation}
\begin{aligned}
 J= &\overbrace{ \nicefrac{\left ( N^{(0)}\mathrm{tr}(\bm{\Sigma}_{(0)}) + N^{(1)}\mathrm{tr}(\bm{\Sigma}_{(1)}) \right )}{\left (N^{(0)}+N^{(1)}\right )}}^\text{inconsistency $\sigma^2$} \\
 &+ \underbrace{(\lVert \bm{\mu}_{(1)}- \bm{\mu}_{(0)}  \rVert _2 - {d})^2}_\text{distortion $\epsilon^2$} 
 + \lambda \underbrace{\lVert \vec{\Theta}- \vec{\Theta}_{\text{n}}  \rVert ^2 }_\text{regularization}. 
\end{aligned}
\label{eq:cost_function}
\end{equation}

\normalsize
To summarize each of the three terms:
\begin{enumerate}
    \item the \textbf{inconsistency} (or variance), denoted $\sigma^2$,  tries to minimize the predicted spread of ball center point that must consistently result in one socket or another; 
    \item the \textbf{distortion} (or bias), denoted $\epsilon^2$, forces the predicted averages to be at the known distance $d$; 
    \item the \textbf{regularization} pushed the optimization to the minimal modification of the nominal model parameters $\vec{\Theta}_{\text{n}}$ (Occam's Razor). This ensures that the optimization converges to the most similar (yet calibrated) set of parameters to the nominal (ideal) model. 
\end{enumerate}
 Since the optimization problem is not linear due to the forward kinematics model, it is common in the literature to initialize the model with the prior nominal model. However, if the nominal model is not provided by the manufacturer, methods in the literature are proposed to retrieve it given the reading from the reading of the black-box controller \cite{faria2019automatic}.

 On the other hand, for amateur projects, when the 3D CAD model is available, different tools can be used to generate the nominal geometric URDF model.
 
 To minimize the volumetric distortion of the method, the only ground truth measure that is exploited in this calibration routine is the distance between the two sockets and the minimal movement of the ball's center point in a given socket. 
 \begin{assumption}
 \label{assumption:distance}
     The uncertainty on the socket distance d is negligible, i.e., $\Delta d \approx 0. $
 \end{assumption}

 This assumption is valid if the manufacturing process has an uncertainty that matches the goal accuracy of the robot. 
 
 Moreover, Fig. \ref{fig:cover} shows two notches on the side of the sockets that are designed to precisely place a caliper and measure the actual post-manufacturing distance, making the Assumption \ref{assumption:distance} to hold. 
 
The proposed minimalist tool only has two sockets, different from other strategies that require three sockets \cite{petrivc2023kinematic} or three spheres \cite{gaudreault2018self, joubair2015kinematic, nadeau2019impedance}. 

 If only one socket is used, the optimization of the parameters may converge to any of the equivalent kinematics that show consistent nullspace prediction but have proportionally bigger (or smaller) link dimensions. On the other hand, by having two sockets at a known distance, the optimization converges to the only set of link dimensions that match the real ones. This ensures that our robot has high volumetric accuracy \cite{alam2022inclusion}. 
We only require three Cartesian quantities to ensure the convergence of the calibration. 
The only advantage of using three spheres (or sockets) is three additional independent quantities (two distances and one consistency), improving numerical stability, reducing sensitivity to measurement noise, and ensuring more robust and accurate parameter estimation in the optimization process. However, in this article, we experimentally validated that the minimal hardware with two sockets is enough for obtaining accurate calibration.
The next section highlights how by recording a set of robot configurations and a few optimization steps, the robot kinematic parameters are updated to reach sub-millimetric precision. Moreover, in the experimental section, we highlight that while training on a single tool position is sufficient, it is not restrictive. If additional datasets are collected from different parts of the workspace, the optimization procedure ensures consistency and minimizes distortion across all of them by simply summing the cost for each tool, described in Eq.~\ref{eq:cost_function}.

\begin{table*}[ht!]
\caption{Mean Absolute Error (MSE) before and after the calibration on Franka, Kuka and Kinova Robots. }
\centering
\begin{tabular}{@{}p{0.12\textwidth}*{4}{L{\dimexpr0.22\textwidth-2\tabcolsep\relax}}@{}}
\toprule
& \multicolumn{2}{c}{\textbf{Training Set}} &
\multicolumn{2}{c}{\textbf{Test Set}} \\
\cmidrule(r{4pt}){2-3} \cmidrule(l){4-5}
& \textbf{Mean Absolute Error (Orig.$\rightarrow$ Opt.}) & \textbf{Removed error} ($\%$)& \textbf{Mean Absolute Error (Orig.$\rightarrow$ Opt.}) & \textbf{Removed error} ($\%$) \\
\midrule
Panda (1)&  $\num{8.28e-03} \rightarrow \num{1.68e-04}$ & 97.97 \% & $\num{7.79e-03} \rightarrow \num{3.47e-04}$ & 94.92 \% \\
Panda (2) & $\num{3.32e-03} \rightarrow \num{1.51e-04}$ & 95.43 \% & $\num{3.26e-03} \rightarrow \num{2.38e-04}$ & 92.69 \% \\
Panda (3) & $\num{8.79e-03} \rightarrow \num{1.84E-04}$ & 97.91 \%  & $\num{1.06e-02} \rightarrow \num{2.62e-04}$ & 97.53 \% \\
FR3 & $\num{8.51e-03 } \rightarrow \num{1.61e-04}$ & 98.11 \% & $\num{8.81e-03 } \rightarrow \num{2.78e-04}$ & 96.84 \% \\
KUKA iiwa 14& $\num{7.23e-04} \rightarrow \num{1.07e-04}$ & 85.23 \% & $\num{ 8.06e-04 } \rightarrow \num{2.68e-04}$ &  66.76 \% \\
Kinova Gen3lite & $\num{4.72e-03} \rightarrow \num{1.52e-03}$ & 67.77 \% & $\num{5.99e-03} \rightarrow \num{2.66e-03}$ &  52.08 \% \\

\bottomrule
\end{tabular}
\label{table:performance}
\end{table*}
\section{Experiments}

In order to validate the proposed tool and optimization methodology, we performed the data recording and optimization for four different Franka Robots (three Panda and one FR3) with 7 DoF, one LBR iiwa 14 R820 with 7 DoF, and two Kinova Gen3Lite with 6 DoF.

The calibrated model was later used in a Cartesian impedance control \cite{albu2003cartesian}
for the Franka Robot to validate the calibrated model performance in high-precision insertion tasks or drawing retracing. 
The insertion and drawing were repeated from different joint configurations, demonstrating the calibrated model's high consistency in insertion and retracing, in contrast to the original nominal model.

Moreover, it also performs the insertion on different known provided offsets on a calibrated table, showing that the distance tolerance matches the accuracy of \SI{0.1}{\milli \meter} of the calibrated table. Volumetric calibration is particularly important when learned (or programmed) skills need to be generalized to different parts of the workspace.
The employed 3-D printing is a commercial entry-level BambuLab A1. This is a very accurate still affordable device. The accuracy is of order of $\SI{0.1}{\milli \meter}$, satisfying Assumption \ref{assumption:distance}. 
% \todo[inline]{Talk about the precision of the printer \url{https://www.techradar.com/pro/bambu-labs-a1-review} 
A video of the experiments is available at: \url{https://www.youtube.com/watch?v=1L7NhbKIDDE}.
\begin{figure}[ht!]

      \centering
    \includegraphics[width=1\linewidth]{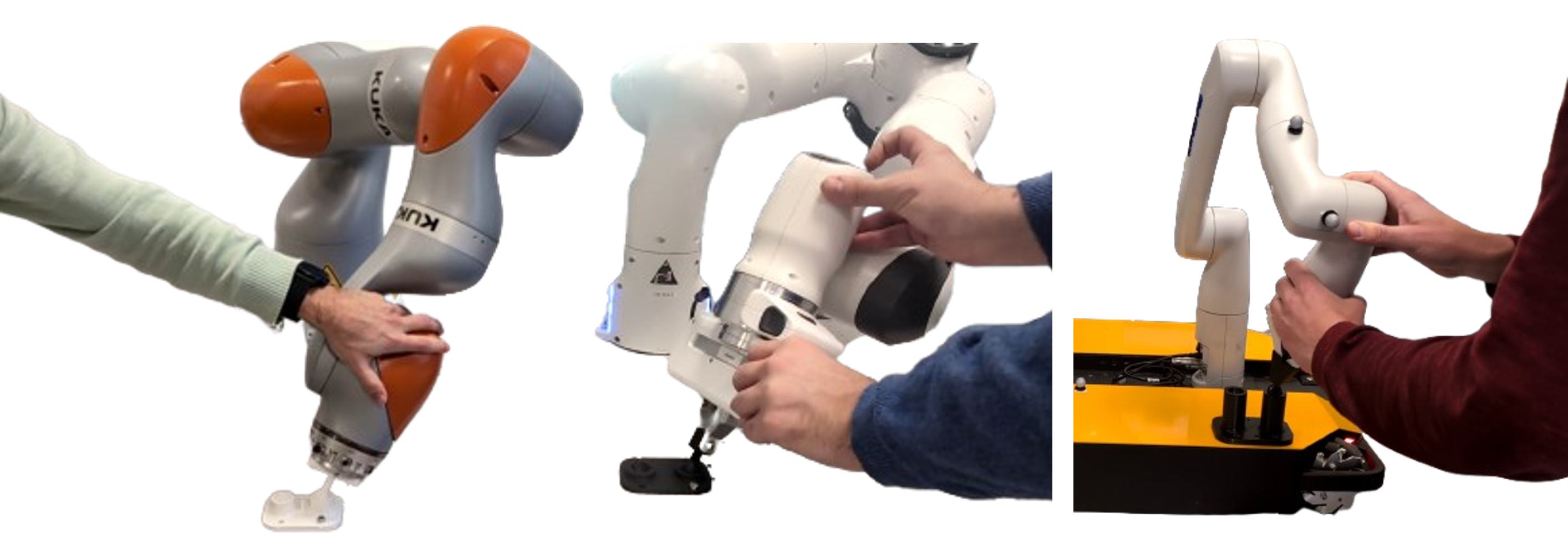}
    \caption{Robots that were calibrated with the MUKCa tool. }
    \label{fig:robots}
    \begin{subfigure}{0.48\linewidth}
    \centering
\includegraphics[width=\linewidth]{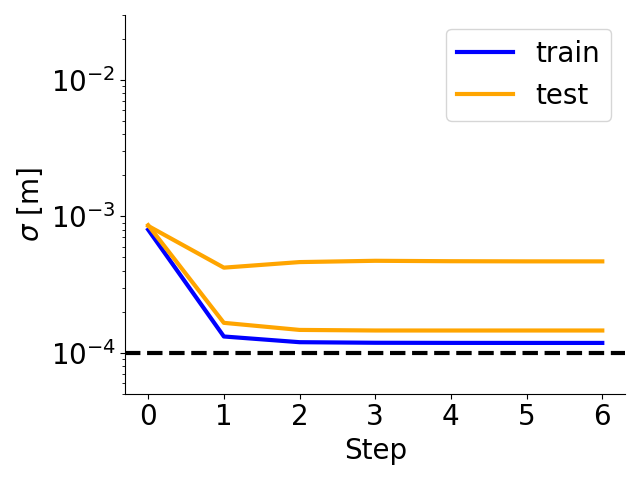}
  \end{subfigure}
  \begin{subfigure}{0.48\linewidth}
    \centering
    \includegraphics[width=\linewidth]{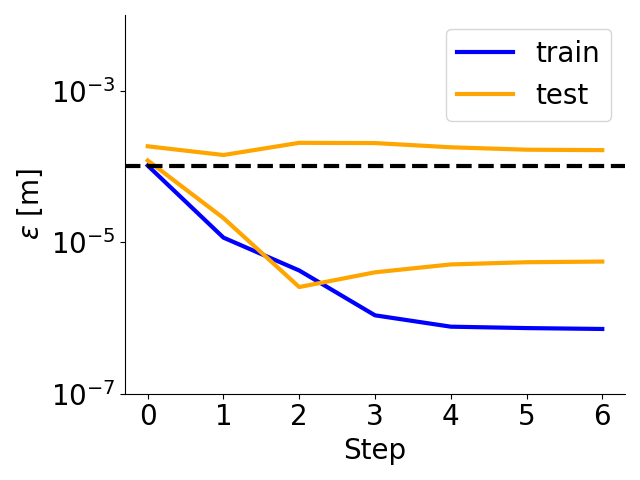}
  \end{subfigure}%
  \caption{Training curves in terms of consistency and distortion KUKA iiwa 14.}
  \label{fig:learning_kuka}
  \begin{subfigure}{0.4\linewidth}
    \centering
    \includegraphics[width=\linewidth]{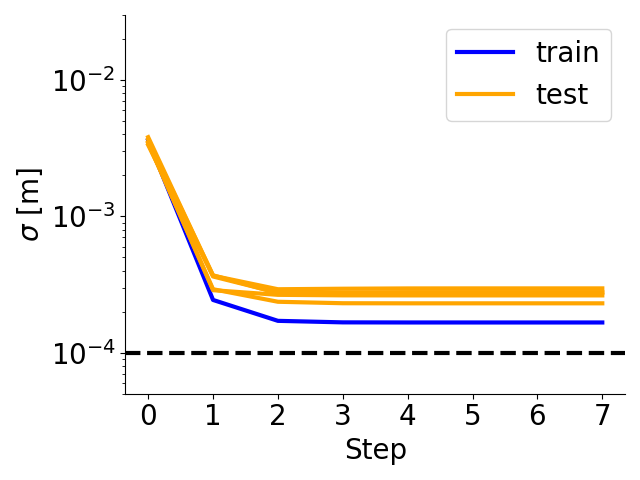}
  \end{subfigure}
  \begin{subfigure}{0.48\linewidth}
    \centering
    \includegraphics[width=\linewidth]{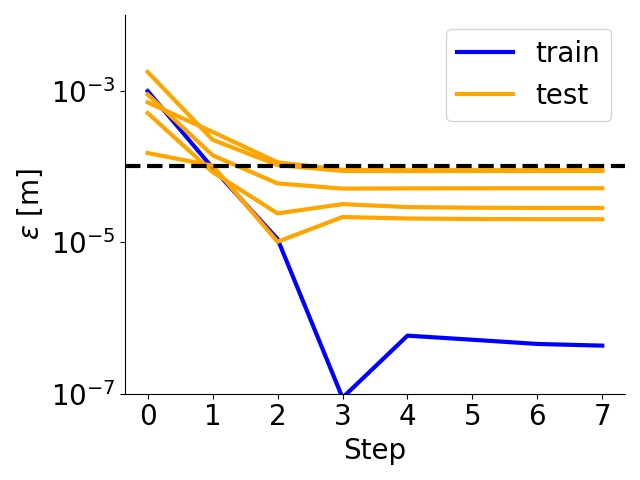}
  \end{subfigure}%
  \caption{Training curves for a Franka``Panda'' (2).}
\label{fig:learning_panda_2}
  \begin{subfigure}{0.48\linewidth}
    \centering
\includegraphics[width=\linewidth]{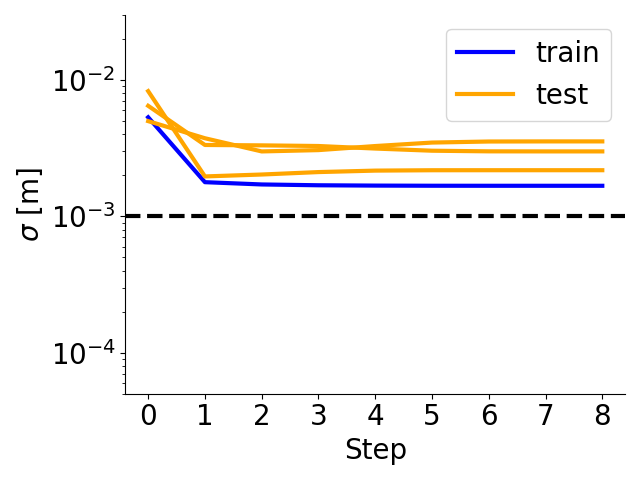}
  \end{subfigure}
  \begin{subfigure}{0.48\linewidth}
    \centering
    \includegraphics[width=\linewidth]{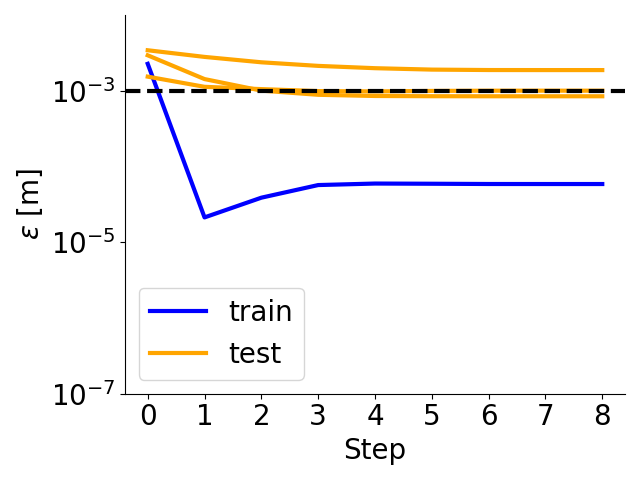}
  \end{subfigure}%
  \caption{Training curves for a Kinova Gen3Lite. } %this is kinova 2
  \label{fig:learning_kiova_2}
\end{figure}
\subsection{Calibrated model identification and validation}
\begin{figure*}[t!]
    \vspace{-5mm}
    \centering
    \includegraphics[width=1\linewidth]{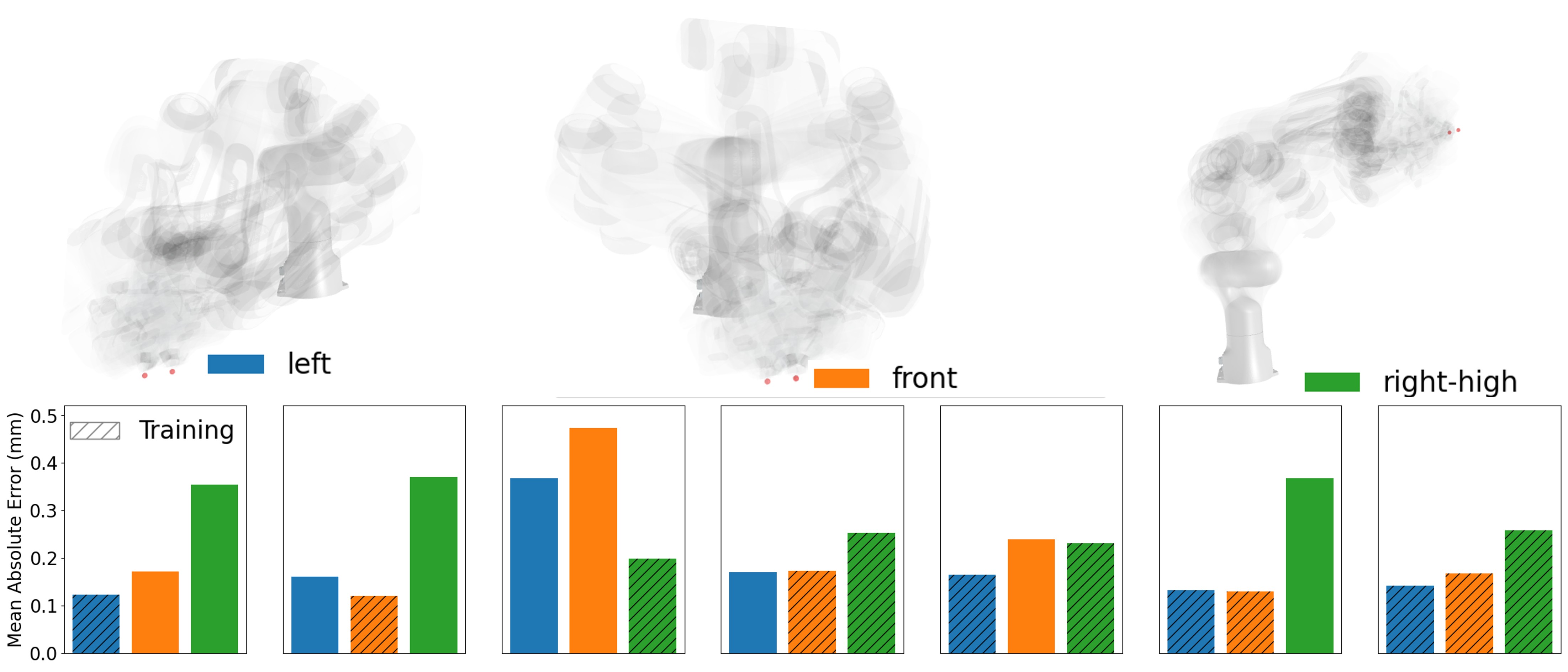}
    \caption{Changing of converging error when training on multiple tool positions. The overlay shows the prediction after training for the tool. The hashed bar implies that the data for that tool was observed during training. }
    \label{fig:bar_plot}
\end{figure*}
For any of the three families of robots to calibrate, we attached the spherical tool at the end-effector and positioned the two-socket tool in front of the robot. Setting the robots in a gravity-compensated and back-drivable modality, allows to easily move the robot in different configurations with the spherical joint in the sockets. The users are asked to use the computer keyboard to add an observation or the buttons on the robots (available only on Franka) with a minimum of thirty samples per socket. One dataset is enough to perform the kinematic calibration, which can take no more than a few minutes to record. However, in our study, we collected many datasets for each robot to test the consistency and distortion of the robot in different parts of the workspace. 
The unconstrained optimization problem described in Sec. \ref{sec:methodology} with a regularization of $\lambda = \num{1e-4}$, is solved using an Interior Point Optimizer \cite{wachter2006implementation}, and if the relative change in the objective function is smaller than $10^{-8}$, the optimization is terminated. The exact first and second-order derivatives of the cost function with respect to the model parameters are obtained using the Automatic Differentiation engine of CasADi \cite{andersson2012casadi}. The cost function is the one reported in Eq.~\eqref{eq:cost_function} that maximizes the consistency and minimizes the distortions while regulating the parameters to be as similar as possible to the nominal model. 

Figure  \ref{fig:learning_panda_2}  shows the optimization curve for a 7-DoF Franka robot. In a few iterations, the optimizer brings the error and the distortion from an order of magnitude of centimeters to submillimeters. This highlights that these robots are potentially very precise but the inaccuracy in their assembly generates a discrepancy between the nominal and the real model. The MUKCa tool closes the gap between the model and reality. Table \ref{table:performance} highlights how the mean absolute error, computed as
$$
\small
   \frac{1}{N}\left ( \sum_{i=1}^{N^{(0)}} \lVert\fk_{\vec{\Theta}}(\bm{q}^{(0)}) -\bm{\mu}^{(0)}\lVert + \sum_{i=1}^{N^{(1)}}  \lVert \fk_{\vec{\Theta}}(\bm{q}^{(1)}) -\bm{\mu}^{(1)} \lVert \right )
$$
converges to values as little as \SI{0.15}{\milli \meter} deleting up to 98 \% of the original error making the robot (almost) as accurate as repeatable. The dashed line depicts the repeatability of the robot.  

Similarly, Fig. \ref{fig:learning_kuka}, shows the optimization curve for the KUKA iiwa 14. Give the more accurate assembly process, makes the original model already with a submillimetric precision ($\approx \SI{0.7}{\milli\meter}$). Nevertheless, by using the calibration tool, the error reduces to \SI{0.107}{\milli \meter}, matching the robot's repeatability.  

However, approaching submillimetric accuracy is not possible for every robot, in particular, if the assumption on negligible joint uncertainty, is not valid due to backlash. This is the case of Fig. \ref{fig:learning_kiova_2} for the Kinova Gen3lite where the calibration converges to an average of \SI{2}{\milli \meter} starting from an accuracy of \SI{5}{\milli \meter}. However, since the estimated repeatability of the robot is around \SI{1}{\milli \meter}, the calibration routine still managed to correct all the geometric inaccuracies. 

From the experiment, we conclude that robots that have harmonic drives, like the Franka and KUKA, result in very precise calibration. On the other hand, when calibrating robots that use planetary gearboxes, like the Kinova, the calibration did not have the same miraculous improvements. We identify the problem to be in the backlash in the joints that is not observable in the encoder measurements. There is an active effort in the robotics community for the parameterization \cite{alam2022inclusion} and the compensation of the backlash \cite{ma2018modeling} that can complement the research of this paper in decreasing the inaccuracy of cheaper robots. 

Although training the robot in a single tool position is shown to be sufficient to reduce error and improve consistency globally, we observe that accuracy slightly degrades for tool positions farther from the training position. This phenomenon is analyzed and quantified in Figure \ref{fig:bar_plot}. We recorded three widely separated datasets for a Franka robot: one in front of the table, one to the extreme left at table height, and one on the right at a higher position attached to the wall. We trained on each position individually and tested on the others, trained on two positions and tested on the remaining one, and finally trained on all positions simultaneously. The results show that training on the front position yields similar accuracy for the left position but worse accuracy for the higher position.
It is worth mentioning that reaching an accuracy error of $\SI{0.3}{\milli \meter}$ for a faraway position is still an impressive result considering the original $\SI{15}{\milli \meter}$ error. In fact, even when training exclusively in the high position, the error does not converge to the same small quantity as when training in the frontal position. The comparisons suggest that the best overall accuracy is obtained by training on the two most distant tool positions in the dataset, or, if only one position is available, prioritizing the frontal positions.
\subsection{Calibrated Insertion tasks}

To validate the improved performance of the calibrated model obtained using the proposed MUKCa procedure, we perform vertical insertion tasks with a peg of diameter of 10 mm attached at the end-effector and holes with different dimensions, from 16 mm down to 10.4 mm. The goal is to validate the \emph{consistency} of the model when performing the insertion task by starting from different nullspace configurations and the \emph{distortion} of the model when offsetting the hole by a known distance on a calibrated table. The experiment was performed on the Franka robot controlled with a Cartesian Impedance controller. 

\begin{figure}
    \centering
    \includegraphics[width=\linewidth]{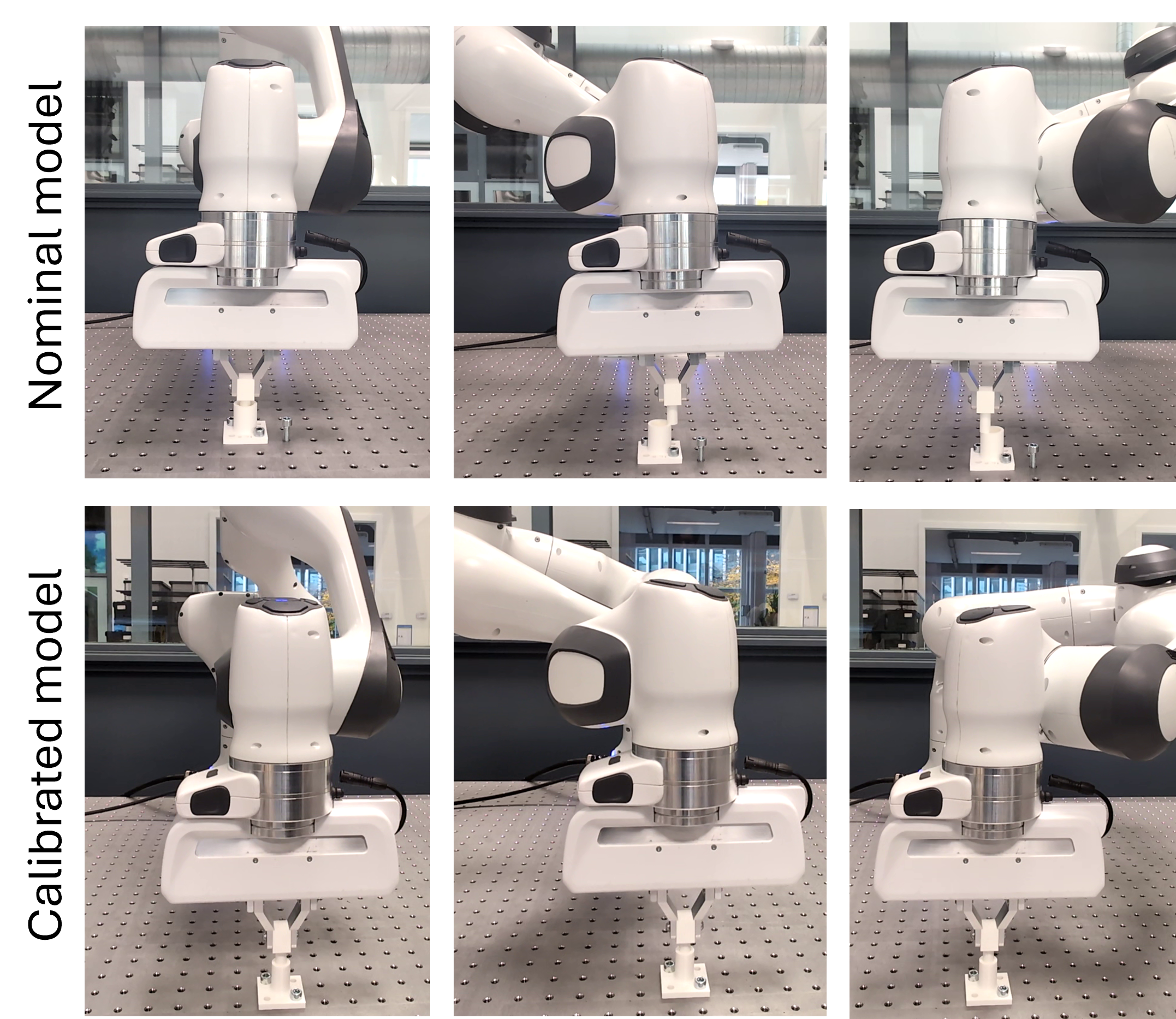}
    \caption{Insertion task with the nominal model or calibrated model. After calibration, the robot successfully inserts the peg (\SI{10}{\milli \meter}) in the hole (\SI{10.4}{\milli \meter}) despite the changes in the initial starting joint configuration. The nominal model even fails with a hole of \SI{16}{\milli \meter}. 
    }
    \label{fig:consistency_calibrated_model}
\end{figure}

Performing consistent insertion tasks with different joint configurations is a necessary feature of redundant manipulators. Figure \ref{fig:consistency_calibrated_model} illustrates the test bench of the performed \emph{consistency} test with a hole of \SI{16}{\milli\meter} of diameter using the nominal model provided by the manufacturer. The goal pose is computed by manually positioning the robot, with the elbow up, in a tight hole and recording the desired Cartesian pose from the forward kinematics. However, when executing the insertion task again with a different starting joint configuration,  the resulting insertion fails despite the high tolerance between the peg and the hole. It is worth explaining that the reason for the failure is not due to controller compliance or manipulability singularities. In fact, we employ high stiffness and compensate for static friction before the insertion, see Appendix \ref{appendix:impedance_control}. Before the insertion, the robot erroneously \emph{believes} that the Cartesian error is very small. This underlines that the issue is the \emph{miscalibration} of the forward kinematics model. 

To prove this, we repeated the experiment using the calibrated forward kinematics and Jacobian model computed using Pinocchio \cite{carpentier2019pinocchio} and the calibrated URDF model. Fig. \ref{fig:consistency_calibrated_model} shows the insertion performance of a hole of 10.4 mm. The insertion was performed 20 times by randomizing the starting joint configuration, and it always resulted in a successful insertion.

Moreover, to validate the volume preservation of the new calibrated model, the hole was moved on a calibrated table of a set of positions 24 other positions positioned on an ``X" shape (6 on each branch). The screws are at a distance of \SI{25}{\milli \meter} in the x and y axis with an accuracy of \SI{0.1}{\milli \meter}. The x-y insertion coordinate is displaced according to the movement of the peg on the table. The experiment resulted in all successful 24 insertions, however, with a lower or higher interaction force due to the inaccuracy of the peg/hole devices, kinematic model, and calibrated table, with a mean force of  $\SI{6}{\N}$  and max force of $\SI{30}{\N}$ and min of $\SI{0.7}{\N}$. 

\begin{figure}[t!]
    \centering
    \includegraphics[width=\linewidth]{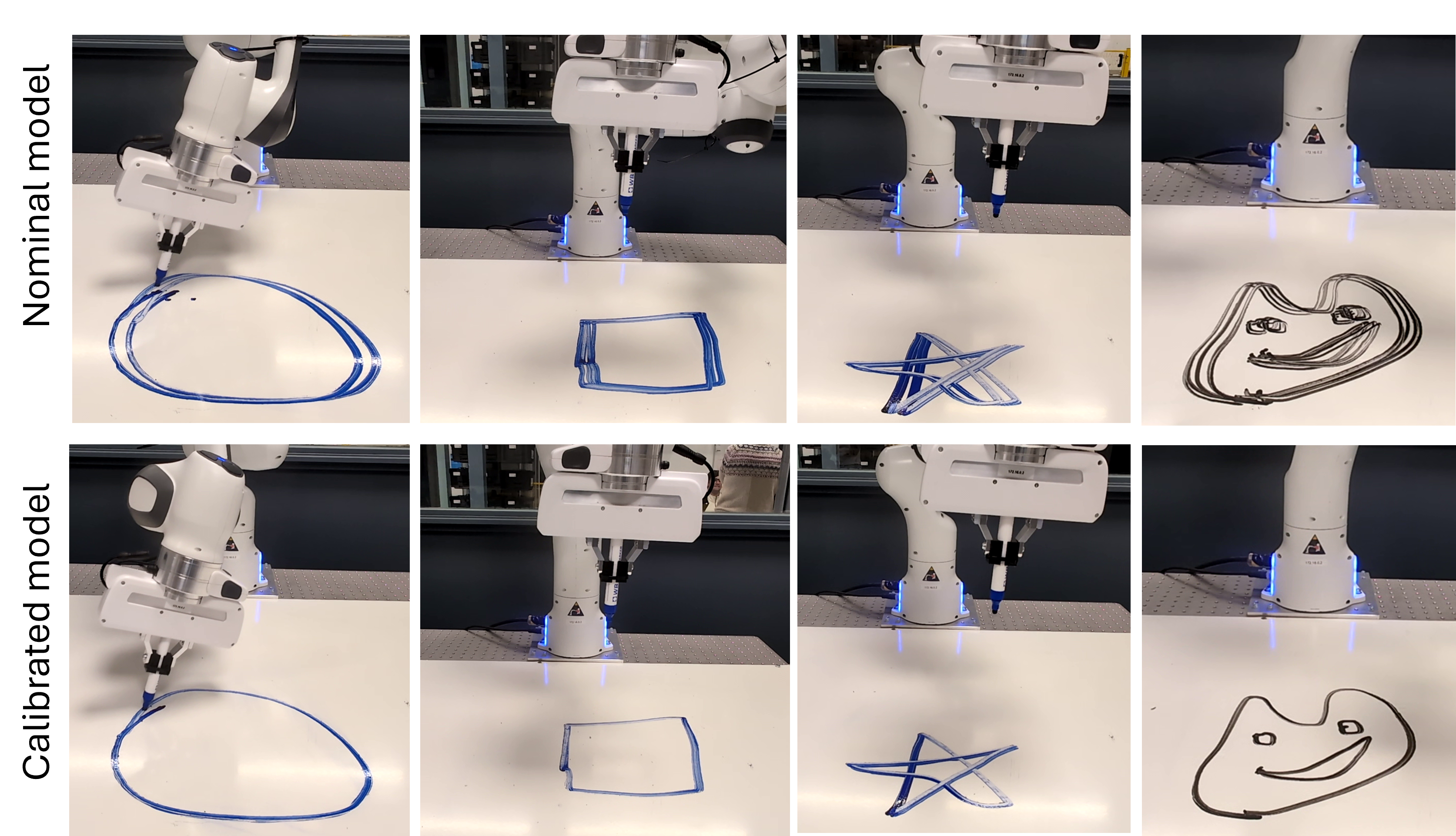}
    \caption{Accuracy of learned drawing motion. By executing the Cartesian motion with different joint configurations, the calibrated models perfectly trace the previous iterations. The nominal model clearly fails on the task. }
    \label{fig:lfd_motions}
\end{figure}

Finally, we also recorded a drawing motion using kinesthetic demonstration and executed it using the calibrated or the nominal model. Figure \ref{fig:lfd_motions}, depicts the attempt of the retracing of the same Cartesian motion starting from a different joint configuration. The figure highlights how the nominal model terribly fails at the talk while the calibrated model consistently retraces the same original drawing.  

\section{Conclusion}
This paper introduces and formalizes a novel calibration tool and optimization routine that only relies on the minimization of the nullspace inconsistency for each of the two sockets and the bias or distortion of the predicted distance between the averages. Despite the simplicity and the affordability of the device that is fully 3D printable, the mean absolute error was reduced from the order of \SI{1}{\centi \meter} centimeter to the order of \SI{0.1}{\milli \meter}. This was also validated by performing high-precision insertion tasks before and after the calibration, showing the improvement from failing insertion tasks with \SI{6}{\milli \meter} tolerance to repeatedly inserting the peg in the hole with only \SI{0.4}{\milli \meter} clearance. 
However, we identify the limitation in calibrating robots that no not guarantee a assume minimal joint measurement error, like the Kinova Gen3lite, due to backlash or sensor noise. We even observed that performing the calibration routine on these robots can result in overfitting, i.e., having a growing error for position further away from the tool.
Future works should focus on applying the methodology to cheaper robots that are not equipped with harmonic drivers or double encoders \cite{mikhel2018advancement} or are not using the torque sensor to estimate and compensate for backlash \cite{giovannitti2022virtual, woodside2024feedforward} or joint deformation \cite{mao2024joint}.  
We foresee exciting challenges and opportunities in the simultaneous calibration of robot geometry, backlash, and joint deformation by only using a simple 3D-printable artifact like in this paper. 

\bibliographystyle{IEEEtran}
\bibliography{reference}

\appendix
\section{Cartesian Impedance Control with Offset Compensator}
\label{appendix:impedance_control}

Impedance control is crucial in robot manipulation because it enables robots to interact with their environments in a flexible, adaptive, and safe manner.
Traditional position or force control methods can be too rigid, resulting in failures or damage when dealing with unpredictable or dynamic environments. 
This control allows robots to modulate the relationship between applied force and the resulting motion, mimicking human-like adaptability in tasks requiring both precision and delicacy, such as assembly, object handling, or interaction with humans.
By dynamically adjusting its stiffness, the robot can control the accuracy in the performing task. 
This adaptability makes impedance control particularly valuable for tasks that involve physical contact, ensuring safer human-robot collaboration and more effective execution of insertion and fine manipulation tasks.

The dynamic equation of a robot manipulator is defined according to
\begin{equation}
    \bm{\mathcal{M}}(\bm{q}) \ddot{\bm{q}}+\bm{\mathcal{C}}(\bm{q},\dot{\bm{q}})+\bm{\mathcal{G}}(\bm{q})= \bm{\tau}_{c}+ \bm{\tau}_{comp}+ \bm{\tau}_{ext} 
\end{equation}
where, in order, from left to right, there are the mass, the Coriolis, and the gravitational term that depend on the joint configuration $\bm{q}$ and, on the right, the torque for the Cartesian (or task) control, the torque that compensates for functions and other unmodelled dynamics and the externally applied torques. 
The task space torque is computed as
\begin{equation}
\label{eq:impedance_control}
    \bm{\tau}_{c}=\bm{{J}}_{\vec{\Theta}}^\top(\bm{\mathcal{K}}\left (\bm{x}_{goal}-\fk_{\vec{\Theta}}(\bm{q}) \right )-\bm{\mathcal{D}}(\bm{{J}}_{\vec{\Theta}}\dot{\bm{q}}))+\bm{\mathcal{C}}(\bm{q},\dot{\bm{q}})+\bm{\mathcal{G}}(\bm{q})
\end{equation}
where the stiffness $\bm{\mathcal{K}}$ and the damping $\bm{\mathcal{D}}$ give the compliant behavior with a critically damped response \cite{albu2003cartesian}. The error from the desired pose $\bm{x}_{goal}$ is computed using the forward kinematics $\fk_{\vec{\Theta}}(\bm{q})$. The projection of the Cartesian forces in joint torques is obtained using the transpose of the geometric Jacobian $\bm{J}_{\vec{\Theta}}$. Both the Jacobian and the forward kinematics depend on the model parameters $\vec{\Theta}$. From Eq. \eqref{eq:impedance_control}, it is evident that relying on a miscalibrated model for control is a recipe for disaster, accumulating errors in the position tracking, the Cartesian velocity estimation, and the force-torque projection.  

Moreover, since we are testing high-precision tasks, no residual error due to the unmodelled static friction in the joints is acceptable. Without compensating the static friction, not even a perfectly calibrated robot can perform a sub-millimetric insertion task because the tracking error would be larger than the insertion tolerance, i.e., $ \epsilon_{trac} = |(\bm{x}_{goal}-\fk_{\vec{\Theta}}(\bm{q}) | > \epsilon_{tol}$, making the even a perfectly calibrated model incapable of performing precise insertion tasks. 

To address this problem, just before the insertion task, an integral compensator is activated for a time interval $\Delta t$, i.e., 
\begin{equation}
    \tau_{comp} = \mathcal{K} \int_0^{\Delta t} \left (\bm{x}_{goal}-\fk_{\vec{\Theta}}(\bm{q}) \right ) dt.
\end{equation}
and set to zero otherwise. The compensator is not activated during the insertion task itself, where there are external forces due to the contact between the peg and the hole. Having the compensator may result in winding up the force that can generate instabilities and/or damage the tool settings.  

For a perfectly compensated controller, 
$   \bm{x}_{goal} -\fk_{\vec{\Theta}}(\bm{q}) \approx 0 $; however, if the forward kinematics is not perfect, i.e., 
$$  \fk_{\vec{\Theta}}(\bm{q}) - \bm{x} =  \epsilon_{cal}(\bm{q}) $$
where the $\epsilon_{cal}(\bm{q})$ is the residual error that we aim to reduce with the calibration procedure. This results in
$$  \bm{x}_{goal} - \bm{x} - \epsilon_{cal}(\bm{q}) \approx 0 $$
showing that the observed Cartesian error is not due to the control or the unmodelled frictions but to the kinematics miscalibration, i.e., 
$$   \underbrace{|\bm{x}_{goal} - \bm{x} |}_{\epsilon_{cart}} \approx |\epsilon_{cal}(\bm{q})|. $$ 

\begin{table}[t!]
    \centering
    \begin{tabular}{c c c c c c c c }
        \hline
        $\mathcal{K}_x $ & $\mathcal{K}_y$ & $\mathcal{K}_z$ & $\mathcal{K}_\alpha$ & $\mathcal{K}_\beta$ & $\mathcal{K}_\gamma$ & $\mathcal{D} $ &$\Delta t [\SI{}{s}]$ \\ 
        \hline
        15000 & 15000 & 2000 & 40 & 40 & 40 & 2$\sqrt{\mathcal{K}}$ & 5\\
        \hline
    \end{tabular}
    \caption{Experimental parameters used in the experiments. The linear stiffness is measured in $\SI{}{N/m}$ and the rotational one in $[\SI{}{Nm/rad}]$}
    \label{tab:parameters}
\end{table}
Table \ref{tab:parameters} reports the used parameters in the experiments for reproducibility.
\end{document}